\documentclass[runningheads]{llncs}

\usepackage{eccv}

\usepackage{eccvabbrv}

\usepackage{graphicx}
\usepackage{booktabs}
\usepackage{wrapfig}
\usepackage{multirow}
\usepackage[ruled,vlined]{algorithm2e}
\usepackage[table]{xcolor}
\definecolor{level1}{gray}{0.95} 
\definecolor{level2}{gray}{0.90} 
\definecolor{level3}{gray}{0.85} 
\usepackage{tcolorbox}
\tcbuselibrary{breakable}
\usepackage{listings}
\usepackage[accsupp]{axessibility}  

\usepackage[pagebackref,breaklinks,colorlinks,citecolor=eccvblue]{hyperref}

\usepackage{orcidlink}

\begin{document}

\title{VADF: Vision-Adaptive Diffusion Policy Framework for Efficient Robotic Manipulation} 

\titlerunning{VADF for Robotic Manipulation}

\author{Xinglei Yu\inst{1} \and
Zhenyang Liu\inst{1,2} \and
Shufeng Nan\inst{1} \and
Simo Wu\inst{1} \and
Yanwei Fu\inst{1,2}}

\authorrunning{X. Yu et al.}

\institute{
Fudan University \and Shanghai Innovation Institute \\
\email{yuxl23@m.fudan.edu.cn,
liuzy24@m.fudan.edu.cn, 
25210980158@m.fudan.edu.cn, 
10300290008@fudan.edu.cn,
yanweifu@fudan.edu.cn}
}

\maketitle

\begin{abstract}
Diffusion policies are becoming mainstream in robotic manipulation but suffer from hard negative class imbalance due to uniform sampling and lack of sample difficulty awareness, leading to slow training convergence and frequent inference timeout failures. We propose VADF (Vision-Adaptive Diffusion Policy Framework), a vision-driven dual-adaptive framework that significantly reduces convergence steps and achieves early success in inference, with model-agnostic design enabling seamless integration into any diffusion policy architecture.
During training, we introduce Adaptive Loss Network (ALN), a lightweight MLP-based loss predictor that quantifies per-step sample difficulty in real time. Guided by hard negative mining, it performs weighted sampling to prioritize high-loss regions, enabling adaptive weight updates and faster convergence.
In inference, we design the Hierarchical Vision Task Segmenter (HVTS), which decomposes high-level task instructions into multi-stage low-level sub-instructions based on visual input. It adaptively segments action sequences into simple and complex subtasks by assigning shorter noise schedules with longer direct execution sequences to simple actions, and longer noise steps with shorter execution sequences to complex ones, thereby dramatically reducing computational overhead and significantly improving the early success rate.
\end{abstract}

\section{Introduction}
\label{sec:intro}

Recent advances in robotic manipulation leverage the synergy between vision-language models (VLMs) and diffusion policies to achieve robust semantic reasoning and multi-modal trajectory generation. Despite their efficacy, current frameworks predominantly rely on a uniform computational allocation strategy that remains agnostic to the inherent \textit{spatio-temporal heterogeneity} of manipulation tasks. This "one-size-fits-all" paradigm fails to account for significant fluctuations in sample difficulty and stage complexity, leading to protracted training convergence and redundant inference overhead.

A primary limitation manifests during optimization, where uniform sampling across diffusion timesteps and data instances fails to distinguish trivial samples from \textit{informative hard negatives}. Although early denoising steps and contact-rich trajectories exert a disproportionate influence on policy robustness, they often receive insufficient optimization focus under a uniform regime, resulting in wasted computation on low-information regions. Furthermore, this lack of compute-awareness extends to deployment, where the rigid application of fixed denoising steps ($N_d$) and action horizons ($N_a$) ignores varying subtask demands. For instance, in a hammering task, preparatory reaching requires far less precision than the critical striking phase. Such uniformity induces a computational mismatch where resources are over-allocated to simple transitions while complex maneuvers are under-served, which manifests as prohibitive latency and degraded success rates.

These challenges reflect a broader absence of \textbf{compute-aware generative control} in robotic imitation learning. While tasks naturally exhibit strong heterogeneity in both instance difficulty and stage-wise requirements, current methods lack the necessary mechanisms for adaptive perception and dynamic resource reallocation.

\begin{figure}[t]
  \centering
  \includegraphics[
    width=1\linewidth,
    trim=5.5cm 9cm 5cm 3cm,
    clip
  ]{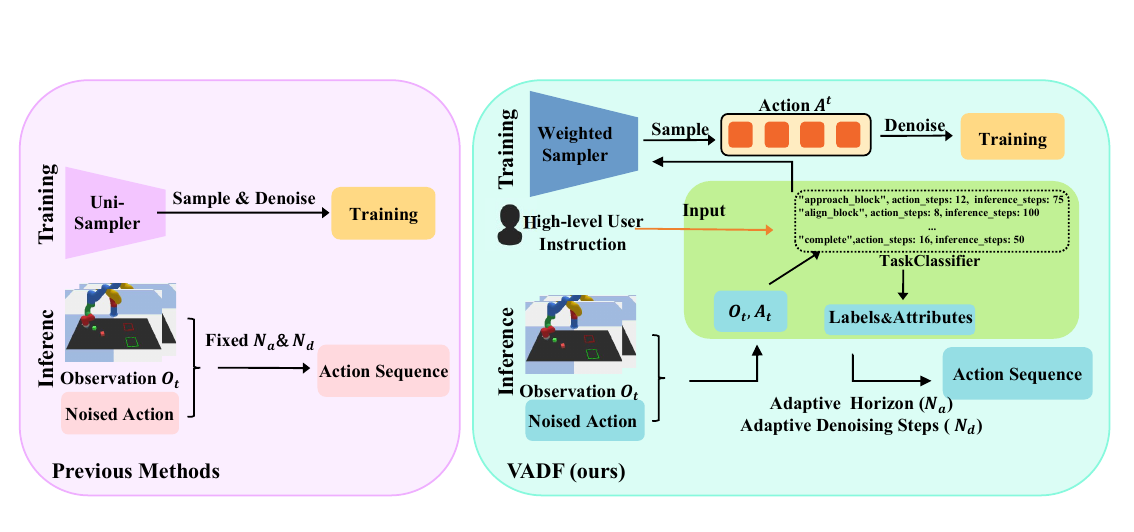}
  \vspace{-0.2in}
  \caption{\textbf{VADF: Vision-Adaptive Diffusion Framework.} VADF introduces a dual-adaptive mechanism for both optimization and deployment. \textbf{ALN :} Accelerates convergence by identifying high-loss regions and prioritizing hard negatives. \textbf{HVTS:} Leverages zero-shot task decomposition to dynamically schedule denoising budgets ($N_d, N_a$) based on stage complexity, ensuring high precision with reduced latency.}
  \label{fig:overview}
   \vspace{-0.15in}
\end{figure}

To address these challenges, we introduce the \textbf{Vision-Adaptive Diffusion Policy Framework}, a model-agnostic and plug-and-play dual-adaptive system (Fig.~\ref{fig:overview}). VADF leverages vision-language understanding to achieve dynamic resource reallocation across both training and inference phases.

Specifically, in the training phase, we propose the \textbf{Adaptive Loss Network}, a lightweight predictor that identifies per-sample reconstruction difficulty in real time. By prioritizing high-loss regions, ALN focuses optimization on hard negatives and critical transitions, significantly accelerating convergence. 

In the inference phase, we present the \textbf{Hierarchical Vision Task Segmenter}, which performs zero-shot task decomposition to segment actions into semantically meaningful stages. Simple stages receive shorter denoising budgets for speed, while complex stages obtain more steps for higher precision, substantially reducing average cost while increasing success rates.

The main contributions of this work are threefold:
\begin{itemize}
    \item We propose VADF, a unified framework that introduces vision-driven adaptive computation into diffusion policies, addressing sample and stage heterogeneity for the first time.
    \item Our dual-module design (ALN and HVTS) is model-agnostic, requiring no modification to underlying architectures, no additional training stages, and no human annotations.
    \item Extensive evaluations across multiple benchmarks (e.g., Push-T, Kitchen, Adroit) demonstrate consistent gains in training speed, inference efficiency, and task performance.
\end{itemize}

By tightly integrating vision-language understanding with diffusion-based generation, VADF establishes a new paradigm for compute-efficient and more interpretable robotic imitation learning, paving the way for diffusion policies to scale to more realistic and complex manipulation scenarios.

\section{Related Work}
\label{sec:related_work}

\noindent \textbf{Sample and Timestep Sampling in Diffusion}
Diffusion training typically assumes uniform importance across noise stages and samples, which is often suboptimal~\cite{kimAdaptiveNonuniformTimestep2025, kingma2023variationaldiffusionmodels}. In image generation, recent works have explored non-uniform timestep sampling~\cite{leeBetaSamplingAll2024, peiOptimalStepsizeDiffusion2025} or instance-aware schedules~\cite{shaoRayFlowInstanceAwareDiffusion2025} to focus on critical perceptual intervals. However, these methods are often handcrafted or limited to vision tasks, overlooking the sample-specific difficulty inherent in robotic trajectories. \textbf{VADF} is the first to bridge this gap in robotic control via the ALN. By replacing static weights with a learned, state-dependent distribution, ALN performs online hard-sample re-weighting, extending importance sampling to diffusion policy learning for the first time.

\noindent \textbf{Acceleration of Diffusion Policies}
Existing acceleration generally falls into three categories: distillation-based approaches~\cite{wangOneStepDiffusionPolicy2024, prasadConsistencyPolicyAccelerated2024} that compress latency but require extensive retraining; single-step models~\cite{liuDiffControlStatefulDiffusionbased2024} that demand architecture modifications; and offline or stochastic scheduling~\cite{wangAdaptiveStochasticCoefficients2025}. While recent works such as Spatial-Temporal Aware Visuomotor Diffusion Policy~\cite{liu2025spatial}, $D^2PPO$~\cite{zouD2PPODiffusionPolicy2025}, and two-step policies~\cite{clemente2025twostepsdiffusionpolicyrobotic} explore improved visuomotor modeling or adaptive denoising strategies, they often lack generalization across diverse task complexities. \textbf{VADF} distinguishes itself via the HVTS, which achieves zero-shot adaptive acceleration. Unlike prior works, HVTS jointly co-adapts both the denoising budget ($N_d$) and action prediction horizon ($N_a$) based on real-time subtask semantics, providing a plug-and-play solution without additional training.

\noindent \textbf{Task Decomposition and Hierarchical Control}
Hierarchical architectures~\cite{zhangHiRTEnhancingRobotic2025, nvidiaGR00TN1Open2025, liHAMSTERHierarchicalAction2024} and Chain-of-Thought (CoT) reasoning~\cite{zhaoCoTVLAVisualChainofThought2025, chenInternVLAM1SpatiallyGuided2025, belkhaleRTHActionHierarchies2024} have shown excellence in long-horizon tasks through semantic planning and spatial decomposition. Related advances in VLA systems and spatial reasoning~\cite{liu2025trivla, liu2026activevla, liu2025reasongrounder, liu2025neural} further highlight the value of structured perception, reasoning, and action for complex embodied decision-making. However, their autoregressive nature often incurs high latency, hindering real-time interaction. \textbf{VADF} overcomes this bottleneck by leveraging a lightweight VLM for stage-wise classification rather than heavy textual reasoning. This enables VADF to achieve fine-grained hierarchical adaptation and efficient motion generation, maintaining the responsiveness required for complex manipulation while filling the gap between high-level planning and low-level dynamic execution.

\noindent \textbf{Positioning of This Work}
Unlike prior acceleration methods that require architectural changes or incur significant inference overhead, \textbf{VADF introduces two complementary, independently ablatable mechanisms}. On the training side, ALN prioritizes high-loss regions for faster convergence; on the inference side, HVTS provides semantic-driven dynamic scheduling of $N_d$ and $N_a$. Together, they enable diffusion policies to adapt to both sample difficulty and task complexity in a fully training-free and model-agnostic manner.

\section{Background: Diffusion Policy}

Diffusion models~\cite{hoDenoisingDiffusionProbabilistic2020}, known for their powerful multimodal modeling and generative capabilities, have become a core paradigm in robotic policy learning. 
Diffusion Policy~\cite{chiDiffusionPolicyVisuomotor2024} was the first to demonstrate that the iterative denoising mechanism of diffusion models outperforms traditional Gaussian policies in high-dimensional continuous control tasks, enabling smoother, more stable, and more diverse action distributions.
Subsequent studies have extended this framework to various domains of robotic manipulation, including trajectory generation~\cite{koLearningActActionless2023}~\cite{zhouRoboDreamerLearningCompositional2024}, grasp planning~\cite{liALDMGraspingDiffusionaidedZeroShot2024}\cite{maDexDiffExtrinsicDexterity2024}, 4D spatiotemporal awareness ~\cite{liuSpatialTemporalAwareVisuomotor2025a}and visual data augmentation for vision-based manipulation~\cite{zhangDiffusionMeetsDAgger2024}, providing new pathways for complex task decomposition, generalizable control, and multimodal perception.

\noindent \textbf{Diffusion Models}. 
Diffusion models are generative models that learn data distributions through a two-stage noising-denoising Markov process. The core idea involves gradually adding Gaussian noise to data samples in the forward process until they become pure noise, then learning to reverse this process to generate new samples from noise. The forward process systematically degrades data through controlled noise injection, while the reverse process uses a trained neural network to predict and remove the added noise step by step.

\noindent \textbf{Diffusion Policy Learning}.
Diffusion Policy adapts the diffusion framework for continuous control tasks by treating action sequences as the data to be generated. Rather than generating images or text, the model learns to generate smooth, multi-step action trajectories conditioned on environmental observations. The key innovation lies in reformulating robot policy learning as a conditional generation problem, where the "data" consists of expert demonstration trajectories.
Its training process involves sampling action sequences from expert demonstrations, adding controlled amounts of noise, and training a neural network to predict and remove this noise while conditioning on visual observations. During inference, the model starts from random noise and iteratively denoises to produce coherent action sequences. A Receding Horizon mechanism is employed where the model predicts multiple future actions but only executes the first few steps before re-observing the environment, balancing long-term planning with environmental responsiveness.

\section{Our Vision Adaptive DiffPolicy Framework (VADF)}

\begin{figure}
  \centering
  \vspace{-0.1in}
  \includegraphics[
    width=0.9\linewidth,
    trim=3.5cm 2cm 4cm 2cm,
    clip
  ]{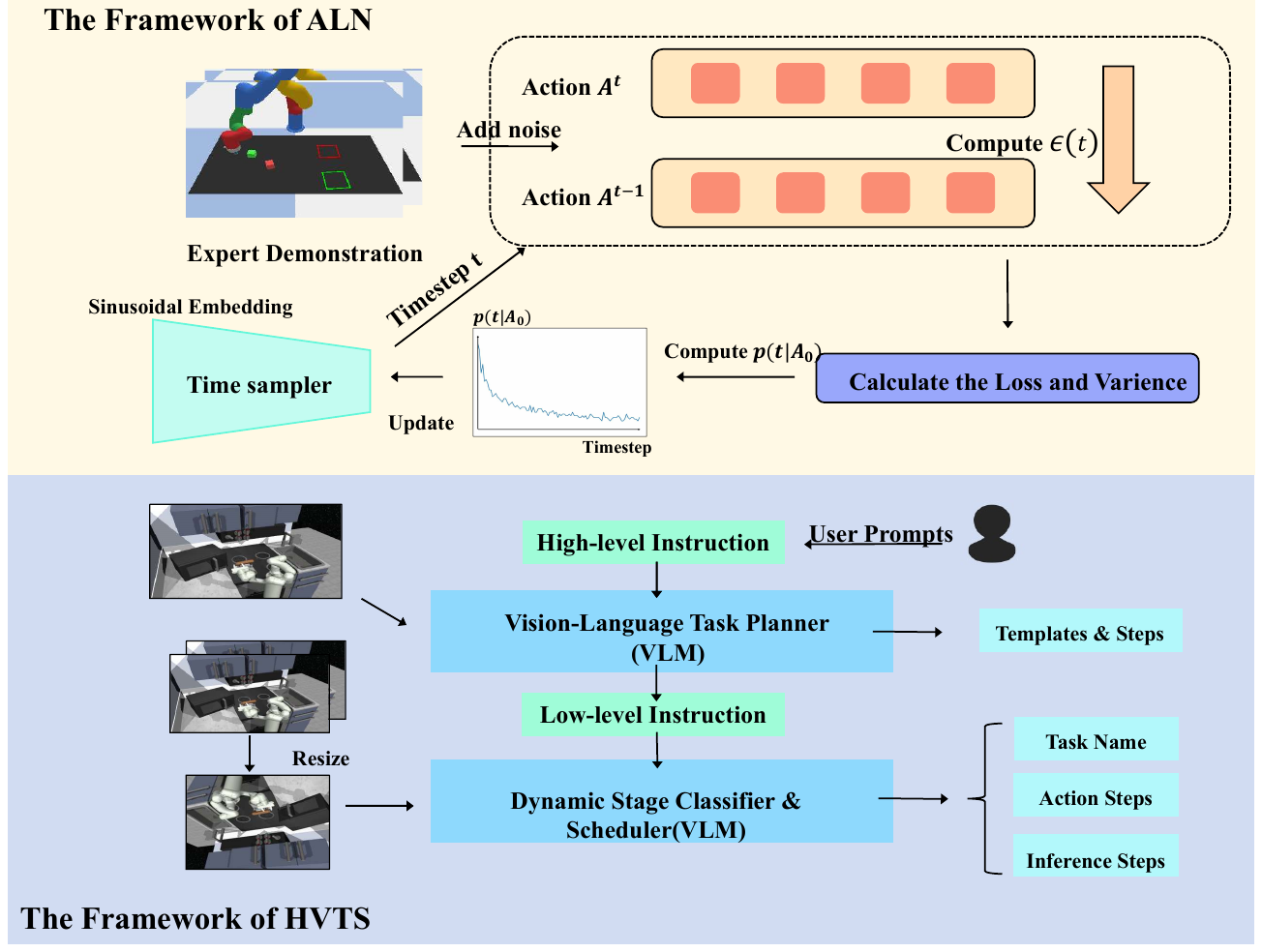}
  \vspace{-0.1in}
  \caption{End-to-end framework of VADF. \textbf{Training phase (ALN):} Expert demonstration action sequences $A^0$ are corrupted with controlled noise to generate $A^{t-1}, A^t$. A learnable time sampler produces temporal encodings $p(t|A_0)$ for the denoising model to predict noise $\epsilon(t)$, with adaptive loss computation and backpropagation enabling efficient policy learning. \textbf{Inference phase (HVTS):} High-level instructions and scene observations are processed by VLM-based task planners and stage classifiers to generate structured task decomposition and dynamic scheduling parameters, enabling adaptive execution across diverse robotic manipulation scenarios.   \label{fig:VADF} }
  \vspace{-0.15in}

\end{figure}

We propose Vision Adaptive DiffPolicy Framework (VADF), a fully adaptive diffusion policy framework. During training, it dynamically optimizes via a learnable timestep sampler and online sample re-weighting. During inference, it achieves zero-shot task decomposition and stage-wise dynamic scheduling using a vision-language model (VLM), jointly controlling denoising steps $N_d$ and action prediction horizon $N_a$. The overall end-to-end architecture, encompassing both the ALN for training and the HVTS for inference, is illustrated in Fig.~\ref{fig:VADF}. The entire system requires no additional training phases, no modifications to the base diffusion model, and no human stage annotations, while simultaneously improving training efficiency, sample efficiency, and inference latency.

\subsection{Mathematical Foundation and Problem Formulation}

To establish the mathematical foundation for our approach, we first formalize the diffusion process and its adaptation to robotic control.

\noindent \textbf{Diffusion Process Formulation}.
In diffusion models, the forward process gradually adds Gaussian noise to real data samples $x_0 \sim p_{\text{data}}(x)$ through a Markov chain:
\begin{equation}
p(x_{k+1} \mid x_k) = \mathcal{N}\!\left(\sqrt{1 - \beta_k}\, x_k,\, \beta_k I \right),
\label{eq:forward_process}
\end{equation}
where $\{\beta_k\}_{k=1}^T$ is a variance schedule, and after $T$ steps, the data degenerates into pure noise $x_T \approx \mathcal{N}(0, I)$.

The reverse process reconstructs clean samples through a noise prediction network $\epsilon_\theta(x_k, k)$. Defining $\alpha_k = 1 - \beta_k$ and $\bar{\alpha}_k = \prod_{i=1}^k \alpha_i$, the reverse sampling step is:
\begin{equation}
x_{k-1} = \frac{1}{\sqrt{\alpha_k}}\Bigg(x_k - \frac{1 - \alpha_k}{\sqrt{1 - \bar{\alpha}_k}}\, \epsilon_\theta(x_k, k)\Bigg) + \sigma_k z,
\label{eq:reverse_process}
\end{equation}
where $z \sim \mathcal{N}(0, I)$ provides additional stochastic noise.

The model is trained by minimizing the mean squared error between the true and predicted noise:
\begin{equation}
\mathcal{L}_{} = \mathbb{E}_{x_0, \epsilon, t}\!\Bigg[ \frac{\beta_t^2}{2 \alpha_t (1 - \bar{\alpha}_t)} \big\| \epsilon - \epsilon_\theta\big(\sqrt{\bar{\alpha}_t}x_0 + \sqrt{1 - \bar{\alpha}_t}\epsilon,t\big) \big\|^2 \Bigg].
\label{eq:ddpm_loss}
\end{equation}

\noindent \textbf{Diffusion Policy Adaptation}.
For robotic control, diffusion policy adapts this framework by treating action sequences as the generation target. From expert demonstration dataset, real action sequences $\mathbf{A}_t^0 \in \mathbb{R}^{T_p \times d_a}$ and corresponding observations $\mathbf{O}_t$ are sampled, and a diffusion step $k \sim \text{Uniform}\{1, \dots, K\}$ is randomly chosen.

The noisy action sequence $\mathbf{A}_t^k$ and visual features $\mathbf{o}_t = f_\phi(\mathbf{O}_t)$ are fed into the noise prediction network $\epsilon_\theta(\mathbf{o}_t, \mathbf{A}_t^k, k)$, and the training loss is computed as:
\begin{equation}
\mathcal{L}_{\text{DP}} = \mathbb{E}_{\mathbf{A}_0, \epsilon, k}\!\Bigg[ \frac{\beta_k^2}{2 \alpha_k (1 - \bar{\alpha}_k)} \big\| \epsilon - \epsilon_\theta(\mathbf{o}_t, \mathbf{A}_t^k, k) \big\|^2 \Bigg].
\label{eq:diffusion_policy_loss}
\end{equation}

The action is updated iteratively during inference as:
\begin{equation}
A_t^{k-1} = \alpha\!\left(A_t^k - \gamma \epsilon_\theta(O_t, A_t^k, k)\right) + \mathcal{N}(0, \sigma^2 I).
\label{eq:action_update}
\end{equation}

\noindent \textbf{Problem Identification}
Standard diffusion policies typically employ a static configuration ($N_d = 100, N_a = 16$) throughout both training and inference, leading to significant inefficiencies:
\begin{itemize}
    \item \textbf{Training Suboptimality}: Uniform timestep sampling ignores varying information gain across noise levels. As Eq.~\ref{eq:diffusion_policy_loss} implies, simple trajectories waste gradient updates on high-noise intervals with minimal policy refinement.
    \item \textbf{Inference Inefficiency}: A fixed denoising budget ($N_d$) and action horizon ($N_a$) are maintained regardless of task difficulty, incurring redundant latency during simple maneuvers while potentially underserving precision-critical segments.
\end{itemize}
To address these gaps, we propose the \textbf{Adaptive Loss Network (ALN)} to prioritize informative timesteps during training, and the \textbf{Hierarchical Vision Task Segmenter (HVTS)} for zero-shot, VLM-driven dynamic scheduling of $N_d$ and $N_a$ during inference.

\subsection{Adaptive Loss Network (ALN)}

To further enhance training efficiency and accelerate hard sample convergence, we introduce two lightweight, parameter-efficient, and independently ablatable adaptive mechanisms into the standard diffusion policy training pipeline.

\noindent \textbf{Learnable Timestep Sampler}.
The theoretical foundation in Equation~\ref{eq:diffusion_policy_loss} contains importance weights that suggest certain timesteps should be prioritized. However, in practice, we typically compute only the vanilla MSE:
\begin{equation}
    \mathcal{L}_{\text{vanilla}} = \mathbb{E}_{\mathbf{A}_0, \epsilon, k} \left[ \|\epsilon - \epsilon_\theta(\mathbf{o}_t, \mathbf{A}_t^k, k)\|^2 \right].
\label{eq:vanilla_mse}
\end{equation}

We can reformulate the standard loss using importance sampling. Since sampling is discrete, if we construct a proposal distribution based on the theoretical weights from Equation~\ref{eq:diffusion_policy_loss}:
\begin{equation}
    q^*(k) \propto w_k = \frac{\beta_k^2}{2\alpha_k (1-\bar{\alpha}_k)}, \quad q^*(k) = \frac{w_k}{\sum_j w_j},
\label{eq:importance_weights}
\end{equation}

and sample $k \sim q^*(k)$, then the vanilla MSE loss in Equation~\ref{eq:vanilla_mse} becomes an unbiased estimator of the original weighted objective in Equation~\ref{eq:diffusion_policy_loss}.

However, $q^*(k)$ is a static distribution derived under the assumption of a near-optimal model. In practice, the per-timestep loss landscape evolves throughout training, and strictly following a fixed $q^*(k)$ may fail to reflect the current difficulty of different timesteps. To better align sampling with the model's evolving error structure, we replace the fixed $q^*(k)$ with a learnable policy $\pi_\phi(k)$ that adapts online and approximates the optimal importance distribution during training.

\noindent \textbf{Algorithm}.
To operationalize the adaptive importance sampling strategy described above, we implement a lightweight timestep sampler as follows.

The timestep $k$ is encoded using sinusoidal positional embeddings and passed through a 3-layer MLP to produce normalized probabilities $\pi_\phi(k)$. 

\begin{algorithm}[!]
\caption{ALN}
\small
\KwIn{
Dataset $\mathcal{D}$; diffusion model $\epsilon_\theta$; 
timestep sampler $\pi_\phi$; total steps $T$; \\
warmup steps $W=500$; entropy coefficient $\lambda=10$; 
reward normalization $\epsilon=10^{-8}$
}
Initialize trajectory weights $\mathbf{w} = \mathbf{1}$

\For{each training iteration}{
    Sample trajectory $i$ using $\mathbf{w}$ via WeightedRandomSampler\;
    
    \If{warmup phase}{
        Sample timestep $k \sim \mathcal{U}(1,T)$\;
    }
    \Else{
        Sample timestep $k \sim \pi_\phi(\cdot)$\;
    }

    Add noise to obtain $\mathbf{A}_t^k$\;
    
    Compute MSE loss 
    $\ell = \|\epsilon - \epsilon_\theta(\mathbf{o}_t, \mathbf{A}_t^k, k)\|^2$\;
    
    Update diffusion model $\theta$ using $\ell$\;
    
    \If{adaptive phase}{
        Normalize loss to obtain reward 
        $r = -(\ell - \mu_\ell)/(\sigma_\ell + \epsilon)$\;
        
        Update sampler via
        $-\!r \log \pi_\phi(k) - \lambda \mathcal{H}(\pi_\phi)$\;
    }
    
    Update trajectory weight 
    $w_i \leftarrow (1-\alpha) w_i + \alpha (r+1)$\;
}
\end{algorithm}

\noindent \textbf{Re-weighting for Hard Negative Mining}.
Beyond timestep-level adaptation, we extend the same importance sampling principle to the trajectory level. 
Specifically, we maintain a lightweight weight vector $\mathbf{w} \in \mathbb{R}^N$ (one per dataset trajectory), enabling data-level importance re-weighting during training. After each batch:
\begin{equation}
    w_i \leftarrow (1 - \alpha) w_i + \alpha (r_i + 1), 
    \quad
    r_i = \frac{l_i - \mu_l}{\sigma_l + 10^{-8}}.
\end{equation}

Here, $\alpha$ follows a cosine annealing schedule, weights are lower-bounded at $10^{-4}$, and globally normalized. Sampling is implemented using PyTorch’s WeightedRandomSampler, increasing the probability of high-loss (hard) trajectories.

Importantly, this mechanism modifies only the sampling distribution while preserving the original diffusion loss formulation, introducing no auxiliary objectives or additional training stages.
This data-level adaptation complements timestep sampling, forming a unified adaptive importance framework across both diffusion steps and training trajectories.

\subsection{Hierarchical Vision Task Segmenter (HVTS)}

Standard diffusion policies employ static inference configurations ($N_d, N_a$), leading to a computational-performance mismatch: simple phases (e.g., reaching) are over-serviced, while precision-critical maneuvers (e.g., insertion) lack sufficient refinement. We propose the \textbf{Hierarchical Vision Task Segmenter (HVTS)}, a zero-shot, training-free plug-in that mirrors the "variable effort" paradigm in human motor control. By dynamically modulating the $(N_a, N_d)$ pair, HVTS achieves high-fidelity execution without the prohibitive latency of uniform high-resolution denoising.

\noindent \textbf{Task Decomposition}. 
HVTS synthesizes visual observations and linguistic instructions via a lightweight VLM to decompose tasks into $K$ semantic segments. Each segment $k$ is assigned a specific computational pair $(N_a^k, N_d^k)$, enriched with pixel-level descriptors and temporal feature priors. This structured decomposition enables VADF to navigate long-horizon manipulations with heightened semantic awareness. \textit{Prompt templates and mapping protocols are detailed in the Supplementary Material.}

\noindent \textbf{Dynamic Scheduling}. 
To translate static decomposition into real-time control, HVTS performs stage identification by cross-referencing live environmental observations $O_t$ (retrieved from a temporal history buffer) against the pre-defined semantic stage library. To preserve temporal coherence and avoid abrupt policy jitter, we implement a probabilistic transition strategy: the VLM evaluates the current scene to output a distribution over the top-$k$ candidate stages, from which the active stage $k$ is sampled to update $(N_a^k, N_d^k)$ dynamically. This mechanism ensures a fluid transition between manipulation phases by assigning accelerated schedules to simple transitions and high-fidelity denoising to precision maneuvers, while maintaining the strict mathematical consistency of the reverse sampling process in Eq.~\ref{eq:reverse_process} and Eq.~\ref{eq:action_update}.

\subsection{Model Architecture}

\textbf{ALN} utilizes a 3-layer MLP (hidden dim: 256) with sinusoidal positional embeddings for the timestep sampler $\pi_\phi$, trained via REINFORCE with entropy regularization. \textbf{HVTS} is implemented using Qwen2-VL-7B-Instruct for both high-level segmentation and real-time stage identification. Notably, VADF is model-agnostic; while we use Qwen2-VL for its strong vision-language alignment, the framework is compatible with any sufficiently capable VLM.

\section{Experiments}


\noindent \textbf{Benchmarks and Datasets} 
We validate VADF across diverse manipulation tasks to ensure robust generalization. Our evaluation encompasses the \textbf{Robomimic} suite~\cite{mandlekar2021matterslearningofflinehuman}, comprising various subtasks such as \textit{Lift}, \textit{Can}, \textit{Push-T}, and \textit{BlockPush}, to challenge the policy's ability to learn from heterogeneous demonstrations in both 2D and 3D spaces.To assess long-horizon reasoning, we employ the \textbf{Kitchen} environment~\cite{guptaRelayPolicyLearning2019} for multi-task sequences, alongside the \textbf{Adroit Task Suite}~\cite{DBLP:journals/corr/abs-1709-10087} for complex 3D dexterous manipulation. These benchmarks support both low-dimensional states and RGB observations, with all experiments conducted across multiple random seeds to ensure statistical rigor.

\noindent \textbf{Expert Demonstrations} 
Expert trajectories are collected at 50 Hz via 6-DoF SpaceMouse teleoperation in MuJoCo~\cite{6386109}. The system records synchronized action vectors, simulator states, and RGB observations. To assess VADF's resilience across data distributions, we adopt the \textit{robomimic} protocol~\cite{mandlekar2021matterslearningofflinehuman} featuring two categories: \textbf{Proficient-Human (ph)} data, containing high-quality single-expert trajectories, and \textbf{Mixed-Human (mh)} data, a heterogeneous multi-operator collection for testing robustness against multimodal, suboptimal demonstrations. Only successful trajectories are retained to ensure foundational behavioral quality.

\noindent \textbf{Baselines} 
We compare against the `vanilla' \textbf{Diffusion Policy}~\cite{chiDiffusionPolicyVisuomotor2024}. To evaluate the efficacy of our dual-adaptive mechanism, we further compare VADF against state-of-the-art DP-based variants, specifically \textbf{DDIM} for accelerated sampling and \textbf{DP3}~\cite{ze3DDiffusionPolicy2024a} for 3D-aware policy learning. These baselines ensure a focused assessment of our method's performance within the diffusion-based imitation learning paradigm under identical backbones and environmental settings.

\noindent \textbf{Implementation Details} VADF is developed atop the official DP codebase. The \textbf{ALN} module utilizes a lightweight 3-layer MLP (hidden dim: 256) with sinusoidal embeddings for learnable timestep sampling, featuring 500 warmup steps and an entropy coefficient $\lambda = 10$. For \textbf{HVTS}, we employ Qwen2-VL-7B for zero-shot decomposition, empirically capping the granularity at $K=5$ stages to balance semantic precision and scheduling stability. The scheduler dynamically co-adapts the action horizon $N_a \in [8, 16]$ and denoising steps $N_d \in [20, 40]$ based on predicted stages. \textit{Detailed sensitivity analyses and the empirical selection criteria for these key hyperparameters ($K, N_a, N_d$) are provided in the subsequent ablation studies.}

Models are trained for 600 to 3,500 epochs (approx. 20k--120k gradient steps) to ensure full convergence across tasks of varying complexity. All experiments are conducted on NVIDIA RTX A6000 GPUs with a batch size of 32. Results are reported as mean $\pm$ standard deviation across 3 random seeds, visualized against gradient steps for consistency.

\begin{table}[t]
\centering
\resizebox{\textwidth}{!}{
\begin{tabular}{l|cc|cc|cc|cc|c|c|c|c|c}
\toprule
\multirow{2}{*}{Method}     
& \multicolumn{2}{c}{Lift} 
& \multicolumn{2}{c}{Can} 
& \multicolumn{2}{c}{Square} 
& \multicolumn{2}{c}{Transport} 
& Push-T 
& ToolHang
& Average 
& Early Succ. (\%) \\
\cmidrule(lr){2-3} 
\cmidrule(lr){4-5} 
\cmidrule(lr){6-7} 
\cmidrule(lr){8-9} 
& ph & mh & ph & mh & ph & mh & ph & mh & ph & ph & & \\
\midrule
IBC 
& 0.79 & 0.15 & 0.00 & 0.01 & 0.00 & 0.00 & 0.00 & 0.00 & 0.00 & 0.00 
& 0.10 & 10.0 \\

DiffusionPolicy-C 
& 1.00 & 1.00 & 1.00 & 0.94 & 0.96 & 0.84 & 0.80 & 0.96 & 0.82 & 0.84 
& 0.92 & 85.4 \\

DiffusionPolicy-T 
& 1.00 & 1.00 & 0.98 & 0.88 & 0.94 & 0.76 & 0.54 & 0.82 & 0.65 & 0.78 
& 0.84 & 80.1 \\

\midrule

\textbf{VADF-C (Ours)} 
& 1.00 & 1.00 & 1.00 & 1.00 & 1.00 & 0.88 
& 0.86 & 0.96 & 0.85 & 0.90
& \textbf{0.95} & \textbf{91.3} \\

\textbf{VADF-T (Ours)} 
& 1.00 & 1.00 & 0.98 & 0.92 & 1.00 & 0.84  
& 0.72 & 0.80 & 0.68 & 0.88 
& 0.88 & 86.8 \\

\bottomrule

\end{tabular}
}
\caption{
\textbf{Behavior Cloning Benchmark (State-Based Policy).}
Success rate (\%) reported from the best-performing checkpoint.
Results are averaged over 3 random seeds, each evaluated on 50 initial conditions per task.
VADF employs HVTS for adaptive inference.
Bold indicates improvement over corresponding Diffusion Policy baseline.
}
\label{tab:state_policy_best}
\vspace{-30pt}
\end{table}

\subsection{Quantitative Comparison}


We show that VADF consistently outperforms strong baselines on two simulation benchmarks.

\noindent \textbf{Behavior Cloning Benchmark}
Table~\ref{tab:state_policy_best} evaluates VADF on the RoboMimic state-based behavior 
cloning benchmark across six manipulation tasks at two proficiency levels 
(proficient-human, ph; mixed-human, mh). VADF-C and VADF-T denote our 
framework applied to the MLP-based and Transformer-based Diffusion Policy 
variants, respectively. VADF consistently matches or improves task success 
rates while substantially boosting early success rate---from 85.4\% to 91.3\% 
for the MLP variant and from 80.1\% to 86.8\% for the Transformer variant---
demonstrating that adaptive inference scheduling accelerates convergence 
to successful trajectories.

\begin{table}
\centering
\caption{Success rates (\%) on the \textbf{Adroit} benchmark (Door, Hammer, and Pen tasks). We compare the original 3D Diffusion Policy (DP3) with \textbf{DP3-VADF}, our proposed improvement of DP3 using the Vision-Adaptive Diffusion Framework (VADF). All methods are trained for 20,000 gradient steps on each task. \label{tab:adroit_dp3} }

\vspace{-0.1in}
\begin{tabular}{lccc}
\toprule
\textbf{Method} &  Door &  Hammer & Pen \\
\midrule
IBC & $0\pm0$ & $0\pm0$ & $9\pm2$ \\
BCRNN & $0\pm0$ & $0\pm0$ & $9\pm3$ \\
\rowcolor{level1} DP3 & $66.7\pm4.7$ & $96.7\pm2.4$ & $65.8\pm9.2$ \\
\rowcolor{level2} \textbf{DP3-VADF} & $\mathbf{76.7\pm6.2}$ & $\mathbf{100\pm0}$ & $\mathbf{76.7\pm2.4}$ \\
\bottomrule
\end{tabular}
\vspace{-0.1in}
\end{table}

\noindent \textbf{Diffusion Policy Benchmark} The benchmark introduced by Chi et al.~\cite{chiDiffusionPolicyVisuomotor2024} spans diverse manipulation skills: Push-T requires precise 2D block positioning, Kitchen demands long-horizon sequential task completion (150–300 steps), and BlockPush involves coordinated multi-object manipulation with obstacle avoidance. As shown in Table~\ref{tab:state_policy_best}, our complete VADF framework consistently outperforms vanilla Diffusion Policy, achieving substantial absolute improvements of 4.5\% on Push-T and 0.6\% on Kitchen. The ablation study reveals that while the ALN component alone (VADF$^{\dagger}$) provides foundational improvements of 1.4\% and 0.2\% respectively, the full framework's superior performance demonstrates the synergistic effect of combining adaptive training optimization with intelligent inference scheduling.

\noindent \textbf{Adroit Simulation} The Adroit benchmark represents one of the most challenging testbeds for vision-based imitation learning, featuring a 24-DoF Shadow Hand performing dexterous manipulation with only 25 expert demonstrations per task, pure image observations, and severe occlusion challenges. Table~\ref{tab:adroit_dp3} demonstrates the plug-and-play nature of VADF: integrating our framework into the state-of-the-art 3D Diffusion Policy (DP3)~\cite{ze3DDiffusionPolicy2024a} yields remarkable improvements of 10.0, 3.3, and 10.9 percentage points on Door, Hammer, and Pen tasks respectively. These results highlight VADF's strong orthogonality, in that it seamlessly enhances existing diffusion architectures while delivering substantial gains under extreme data scarcity and heavy visual occlusion.

\begin{figure}[t]
\centering
\begin{tabular}{cc}
\begin{tabular}{c}
\includegraphics[width=0.2\paperwidth]{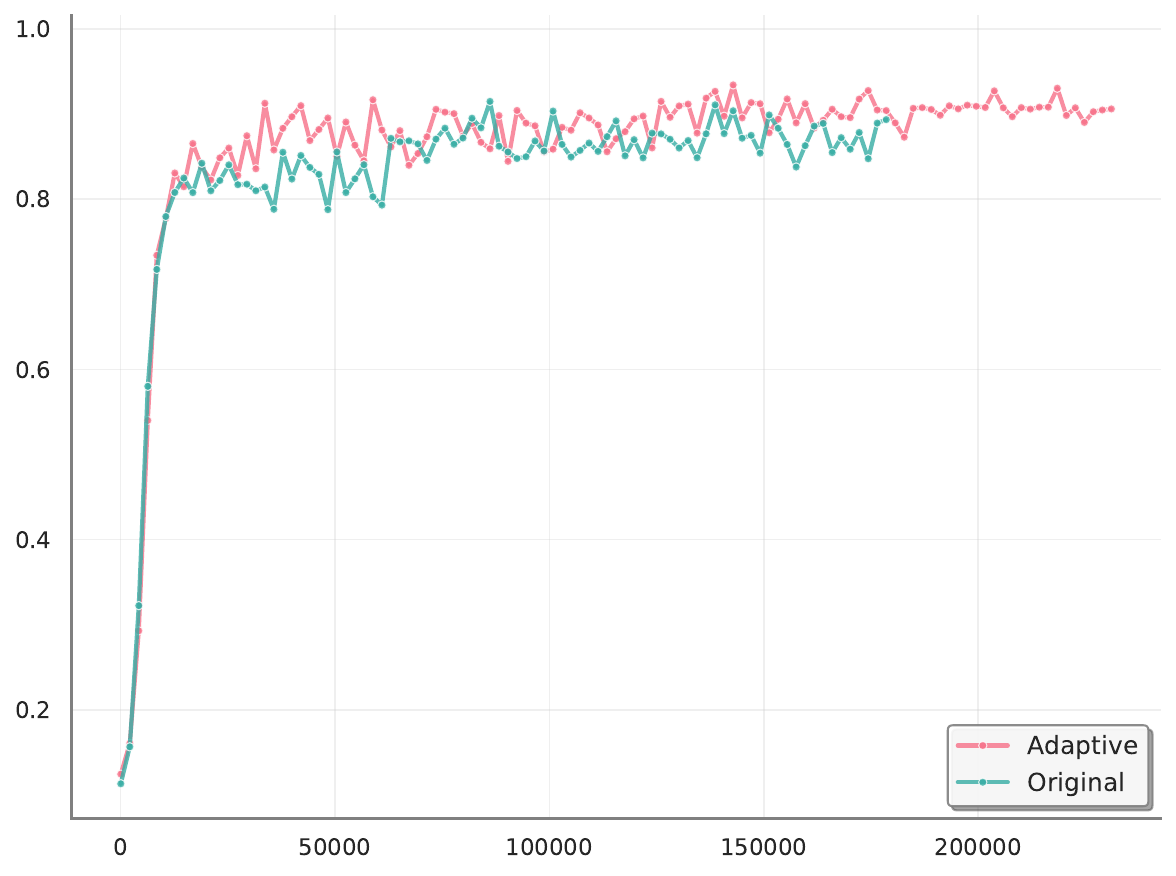}
\end{tabular}
&
\begin{tabular}{c}
\includegraphics[width=0.2\paperwidth]{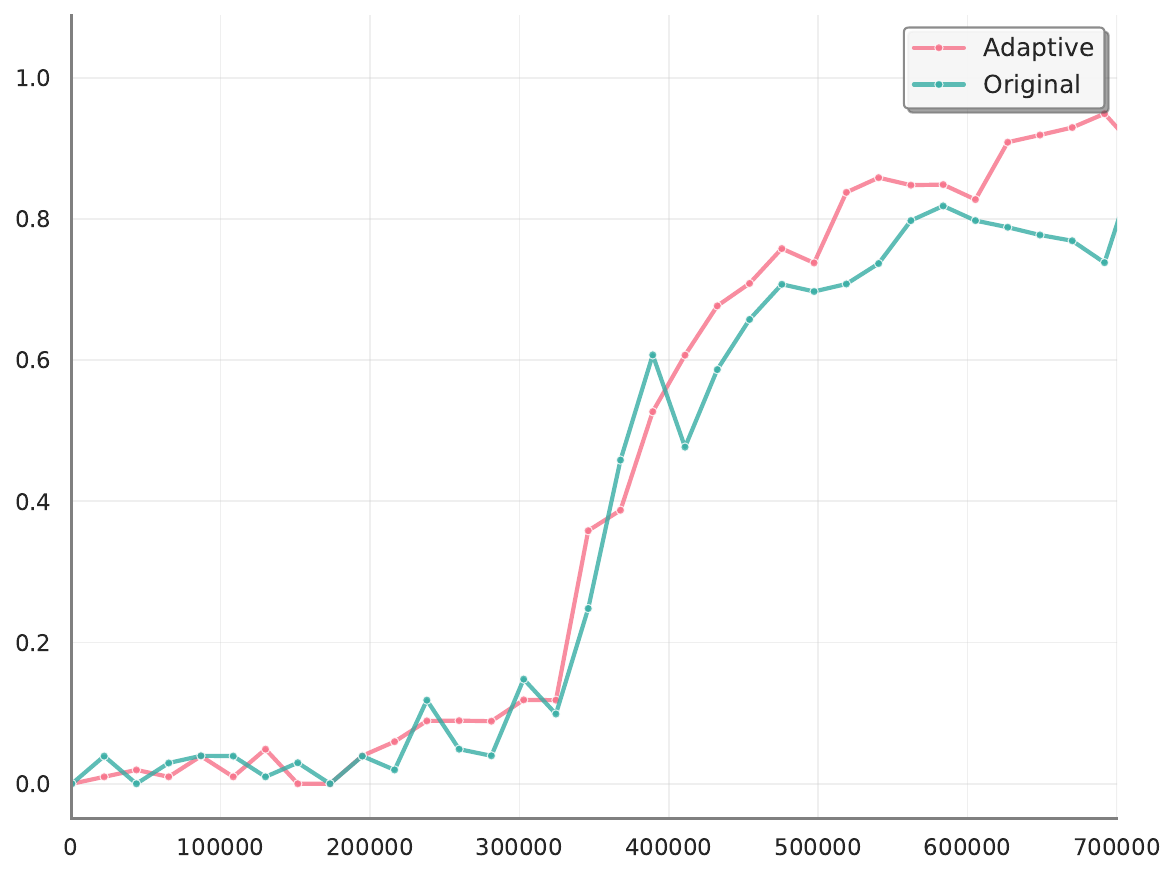}
\end{tabular}
\\
(a) Push-T & (b) BlockPush
\end{tabular}
\caption{Learning dynamics: test scores vs. training steps on low-dimensional tasks. The red curves represent our VADF framework, while the green curves denote Vanilla DP. VADF demonstrates faster convergence.}
\label{fig:learning_dynamics}
\end{figure}

\noindent \textbf{Efficiency Analysis} 
VADF achieves superior efficiency during both training and inference. As illustrated in Fig.~\ref{fig:learning_dynamics}, the ALN module significantly accelerates convergence, reaching peak success rates with substantially fewer gradient steps than vanilla DP while reducing variance. Furthermore, our adaptive scheduling consistently reduces inference latency without degrading performance. As shown in Table~\ref{tab:latency-enhanced}, VADF achieves up to a $2.46\times$ speedup when integrated with DDIM under identical hardware settings, validating its capability to balance high-fidelity execution with real-time computational constraints.

\begin{table}[t]
\centering
\small
\begin{tabular}{l|c|c|c|c|c|c}
    \toprule
    \textbf{Policy}     & \textbf{Succ (\%)} & \textbf{Lat (ms)} & \textbf{Steps} & \textbf{Total Lat (ms)} & \textbf{Freq (Hz)} & \textbf{Speedup} \\
    \midrule
    DDPM & 82.0 & 159.4 & 50 & 7968 & 6.3 & $1.00\times$ \\
    DDPM + VADF & 82.3 & 118.0 & 37  & 4366 & 8.5 & $1.35\times$ \\
    DDIM & 82.7 & 79.8  & 25 & 1995 & 12.5 & $1.99\times$ \\
    DDIM + VADF & 83.3 & 64.8 & 18.5& 1199 & 15.4 & $2.46\times$ \\
    \bottomrule
\end{tabular}
\caption{Inference efficiency comparison on Push-T. Success rate (\%), per-step latency (ms), speedup, and additional efficiency metrics.}
\label{tab:latency-enhanced}

\end{table}

\subsection{Qualitative Comparison}

\begin{figure}[t]
\centering
\includegraphics[width=0.3\paperwidth, trim=3cm 2cm 4cm 3cm, clip]{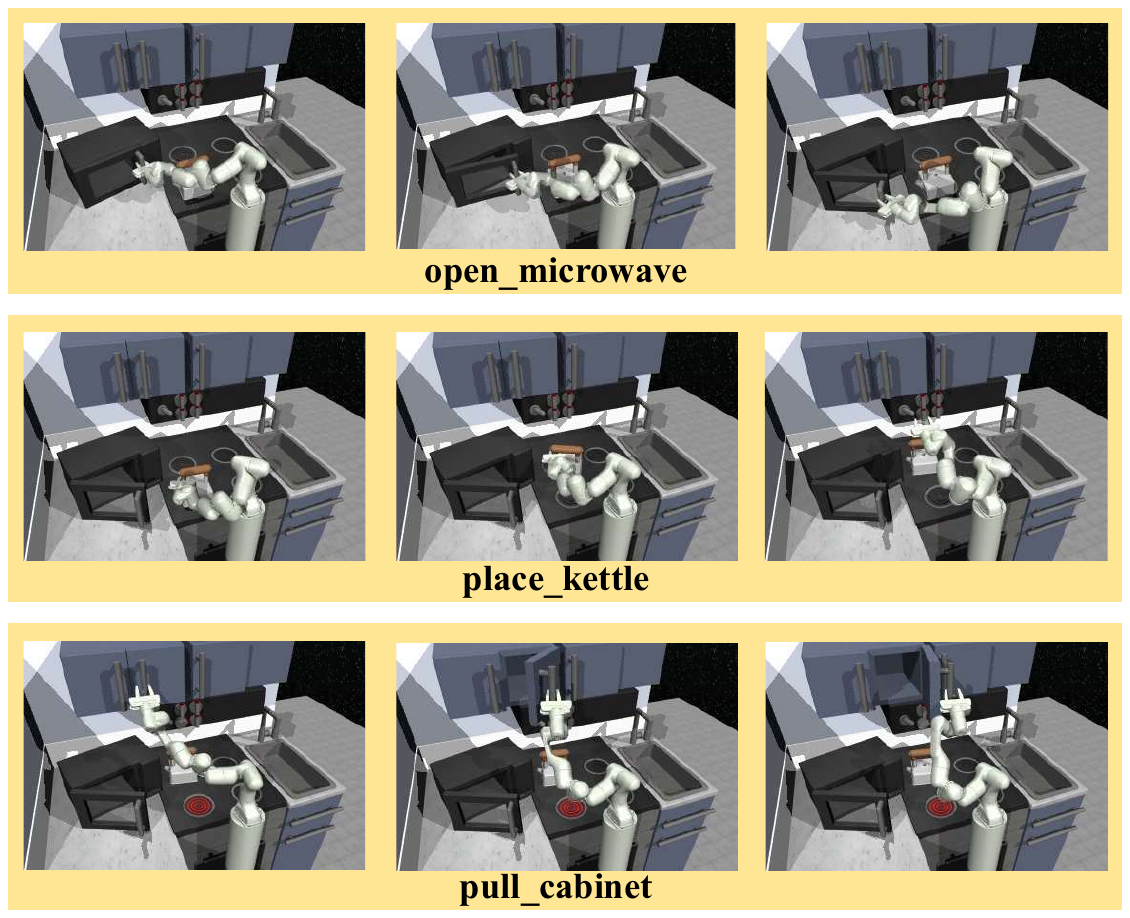}
\caption{Task segmentation and stage recognition results in HVTS for multi-stage tasks. From top to bottom: \textit{open\_microwave}, \textit{place\_kettle}, and \textit{pull\_cabinet}. All outputs are generated online during test rollouts.}
\label{fig:hvts_results}
\end{figure}

To further elucidate the adaptive planning and execution capabilities of the HVTS framework, we conduct a qualitative analysis of its decision pipeline on \textit{Push-T} and \textit{Adroit-Pen} tasks using held-out test episodes. All results are obtained from the final trained policy, with VLM intermediate outputs logged in real-time.

In the \textbf{Push-T} task, given the high-level instruction ``Push the T-block into the green target area," the VLM-based planner first generates a structured template comprising five sequential stages: \textit{approach}, \textit{align}, \textit{push}, \textit{reach}, and \textit{complete}. During execution, the Dynamic Stage Classifier monitors real-time observations to modulate the inference budget. For instance, in the \texttt{align block} stage, where precision is paramount, the VLM detects edge-to-edge alignment and dynamically allocates a higher denoising budget ($N_d$). Conversely, during the \texttt{approach block} phase, the scheduler accelerates inference by reducing steps. In contrast, the baseline without dynamic scheduling often suffers from premature pushing during the alignment phase, leading to accumulated errors and task failure.

In the \textbf{Adroit-Pen} task, characterized by high-dimensional dexterity and severe self-occlusion, the HVTS decomposes the instruction Rotate the pen tip upward" into a three-stage strategy: \textit{initial grasp}, \textit{rotation}, and \textit{stabilization}. Despite complex visual conditions, the Stage Classifier accurately identifies the rotation in progress" stage and maintains stable finger-level control via synchronized step-level scheduling.

As illustrated in Figure~\ref{fig:hvts_results}(c), these results demonstrate that the HVTS framework enables reliable semantic-level task decomposition and vision-driven stage transitions. This fine-grained scheduling ensures that the policy remains both computationally efficient and robust under complex, long-horizon multimodal robotic tasks.

\subsection{Ablation Study}

To validate the effectiveness of each component within the VADF framework, we conduct systematic ablation studies. Table~\ref{tab:dp_ablation} presents comparative results between the full VADF framework and its variants.

VADF$^{\dagger}$ denotes the variant with the vision-language mining mechanism (i.e., HVTS component) removed, retaining only the ALN. The experimental results demonstrate:

\begin{table} \small
    \centering
    \begin{tabular}{lcc|cc}
    \toprule
    \textbf{Method} & \multicolumn{2}{c|}{\textbf{Components}} & \multicolumn{2}{c}{\textbf{Tasks}} \\
    \cmidrule(lr){2-3} \cmidrule(lr){4-5}
    & \textbf{ALN} & \textbf{HVTS} & \textbf{Push-T} & \textbf{Kitchen} \\
    \midrule
    DP & $\times$ & $\times$ & $81.6\pm1.8$ & $99.4\pm0.2$ \\
    VADF$^{\dagger}$ & \checkmark & $\times$ & $83.0\pm0.8$ & $99.8\pm0.1$ \\
    \textbf{VADF} & \checkmark & \checkmark & $\mathbf{86.1\pm1.3}$ & $\mathbf{99.9\pm0.1}$ \\
    \bottomrule
    \end{tabular}
    \caption{\textbf{Ablation Study.} Success rates (\%) on \textbf{Push-T} and \textbf{Kitchen}. 
    We compare our VADF against baseline and ablations. 
    $\checkmark/\times$ indicate component presence/absence. 
    ``VADF$^{\dagger}$'' denotes the variant without vision-language mining. 
    All methods are trained for 3,000 epochs on each task.     \label{tab:dp_ablation} }
\vspace{-0.15in}
\end{table}

On the Push-T task, VADF$^{\dagger}$ achieves a 1.4\% improvement over the original DP method (from 81.6\% to 83.0\%), while the complete VADF framework realizes a substantial 4.5\% improvement (86.1\%). This indicates that the ALN component provides foundational improvements for simple 2D tasks, while HVTS's vision-guided dynamic scheduling further enhances performance.

On the Kitchen task, VADF$^{\dagger}$ yields only a marginal 0.4\% improvement, whereas the full VADF achieves a 0.5\% gain. Despite the small absolute values, this demonstrates the importance of vision-language decomposition for handling complex task sequences in long-horizon multimodal scenarios.

The ablation analysis reveals that while ALN provides consistent but modest improvements across tasks, HVTS is essential for complex manipulation scenarios. The synergistic effect of both components in the full VADF framework demonstrates the complementary nature of adaptive training optimization and intelligent inference scheduling.

\noindent \textbf{Sensitivity to Scheduling Ranges}
Table~\ref{tab:sensitivity_combined} reports task success rate and inference latency (ms) across different combinations of action-step and inference-step ranges. The configuration with action steps in $[8, 16]$ and inference steps in $[20, 40]$ achieves the best trade-off: it preserves high success rates while yielding the lowest latency. Widening either range does not consistently improve accuracy and generally increases latency, so we adopt this as the default setting.

\begin{table}
\centering

\begin{minipage}{.62\textwidth}
    \centering
    \small
    \begin{tabular}{l|c|c|c|c}
        \toprule
        $N_d \backslash N_a$ & 8--12 & 8--16 & 8--20 & 8--24 \\
        \midrule
        20--40   & 84.4 (80.4) & 85.8 (55.3) & 80.3 (56.7) & 84.6 (69.0) \\
        20--60   & 84.4 (80.2) & \cellcolor{gray!20} 88.3 (65.2) & 82.3 (72.8) & 86.7 (73.2) \\
        20--80   & 88.7 (91.5) & 76.1 (58.3) & 83.1 (82.6) & 89.1 (102.2) \\
        20--100  & 83.6 (114.4) & 78.9 (74.8) & 77.5 (85.4) & 83.5 (109.5) \\
        \bottomrule
    \end{tabular}
\end{minipage}
\hfill
\begin{minipage}{.35\textwidth}
    \centering
    \small
    \begin{tabular}{c|c|c}
        \toprule
        $K$ & Succ. (\%) & Lat. (ms) \\
        \midrule
        2 & 74.0 & 82.3 \\
        3 & 84.8 & 95.5 \\
        \textbf{5} & \cellcolor{gray!20} \textbf{88.3} & \textbf{65.2} \\
        7 & 83.9 & 74.5 \\
        10 & 84.1 & 105.3 \\
        \bottomrule
    \end{tabular}
\end{minipage}
\caption{Sensitivity analysis on Push-T. Each cell reports \textbf{success rate (\%)} and \textbf{inference latency (ms)}. (Left) Joint sensitivity of action-step ($N_a$) and diffusion-step ($N_d$) ranges. (Right) Impact of the number of semantic stages ($K$) under fixed ranges ($N_a \in [8, 16], N_d \in [20, 60]$).}
\label{tab:sensitivity_combined}
\end{table}

\begin{figure}[t]
  \centering
  \includegraphics[
    width=1\linewidth,
    trim=1cm 8cm 1cm 8cm,
    clip
  ]{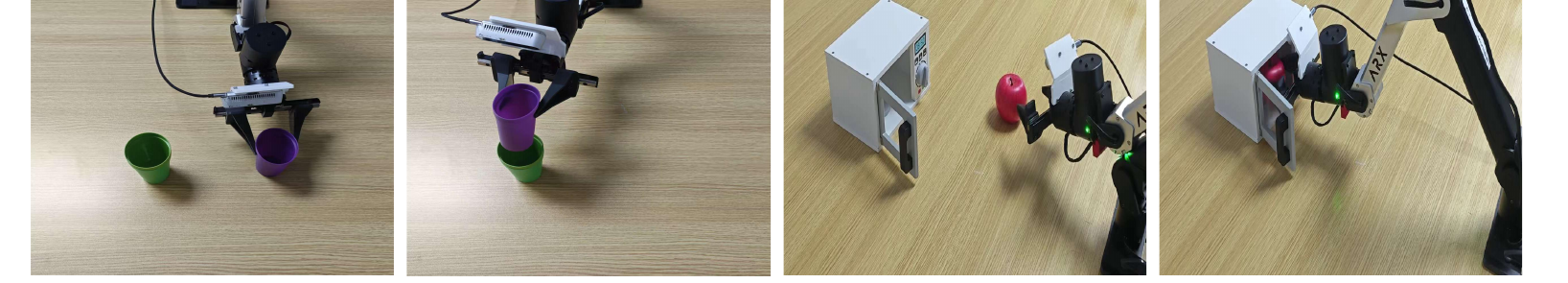}
  \vspace{-0.2in}
  \caption{Real-world inference snapshots on the ARX5 robotic arm. Sequential frames (from left to right) illustrate the robot performing a manipulation task guided by VADF.}
  \label{fig:realrobot}
   \vspace{-0.15in}
\end{figure}

\noindent \textbf{Real-world Validation} 
We further evaluate the cross-domain transferability of VADF on a physical \textbf{ARX5} robotic arm. As illustrated in Fig.~\ref{fig:realrobot}, the robot successfully executes the task through a sequence of continuous reasoning and execution steps. Our results demonstrate that the HVTS module effectively translates semantic task logic into robust physical execution under varying real-world conditions. Detailed experimental configurations, task specifications are provided in the \textbf{Supplementary Material}.

\section{Conclusion}

We propose VADF, a vision-adaptive diffusion policy framework that enhances training efficiency through learnable timestep sampling and hard negative mining, while enabling zero-shot inference adaptation via vision-language guided task decomposition. Experimental results demonstrate consistent improvements across diverse manipulation benchmarks. The framework's plug-and-play design requires no additional training or annotations, making it broadly applicable to existing diffusion policies. 

Despite its efficacy, VADF opens several avenues for future work. First, the current \textbf{HVTS module relies on the zero-shot reasoning of VLMs} for task segmentation; while this facilitates ease of deployment, its reliability is inherently linked to the underlying perception model's performance. Second, although the core philosophy of VADF, \textbf{dynamic allocation of computational resources based on visual task complexity}, is inherently model-agnostic, this study primarily validates its impact on diffusion-based policies. We believe this "adaptive-compute" paradigm holds significant potential for a broader range of generative robotic models, and extending it to other policy architectures remains a promising direction. In summary, VADF establishes a new paradigm for efficient and interpretable robotic imitation learning.

\clearpage

%
%
\bibliographystyle{splncs04}
\bibliography{main}

\clearpage
\appendix

\begin{center}
    {\LARGE \bfseries Supplementary Material for VADF}
\end{center}

\vspace{1em}

In this supplementary material, we provide additional technical details and experimental results that were omitted from the main paper due to space constraints. Specifically, Section \ref{sec:real_world} details our real-world robotic setup and additional qualitative results. Section \ref{sec:hvts_details} expands on the implementation of the Hierarchical Vision-Tactile Switching (HVTS) framework.

\section{Real-world Experiments}
\label{sec:real_world}

\subsection{Robot Hardware Setup}

The real-world experiments are conducted on a \textbf{ARX5} robotic manipulator developed by \textbf{Ark Infinite}, equipped with the default parallel gripper provided with the robot.

\noindent \textbf{Visual Sensors.}
Two RGB cameras are used for visual observation. A wrist-mounted camera is attached to the robot end-effector to provide a hand-eye view of the manipulation region. In addition, a fixed front-facing camera provides a global view of the workspace. Both cameras are \textbf{Intel RealSense D435} sensors.

\noindent \textbf{Computation Platform.}
All real-world policy inference is executed on a workstation equipped with an \textbf{NVIDIA RTX 5880 Ada GPU}. Teleoperation data collection and robot control are executed at a control frequency of \textbf{500 Hz}.

\subsection{Task Specifications}
For each real-world task, we collected 50 expert demonstrations using teleoperation. The baseline policy is \textbf{Diffusion Policy}. We compare the vanilla DP model with our proposed \textbf{DP+VADF} variant. Both models are trained for \textbf{600 epochs} under identical training settings.

\begin{table}[h]
\centering
\begin{tabular}{c|c}
\toprule
\textbf{Task ID} & \textbf{Task} \\
\midrule
Task A & Stack the purple cup onto the green cup. \\

Task B & Place the apple into the microwave. \\

Task C & Insert the cup into the paper slot. \\

\bottomrule
\end{tabular}
\caption{Real-world manipulation tasks evaluated on the ARX5 platform. Each task is repeated for 15 trials.}
\end{table}

During evaluation, each task is executed for \textbf{15 trials}. The primary evaluation metric is the \textbf{task success rate}. For each trial, task completion is determined based on the final manipulation outcome (e.g., correct placement or successful stacking). The overall success rate is computed as the proportion of successful trials across all runs.

\begin{table}[h]
\centering
\label{tab:real_world_results}
\begin{tabular}{lccc c}
\toprule
\textbf{Method} & \textbf{Task A} & \textbf{Task B} & \textbf{Task C} &  \textbf{Avg.} \\
\midrule
DP & 26.7 & 53.3 & 53.3 & 44.3 \\
DP+VADF & 66.7 & 73.3 & 86.7 & 75.6 \\
DP3 & 66.7 & 66.7 & 60.0 & 64.5 \\
DP3+VADF & \textbf{86.7} & \textbf{80.0} & \textbf{86.7} & \textbf{84.5} \\
\bottomrule
\end{tabular}
\caption{Real-world success rate (\%) on three manipulation tasks.}
\end{table}

This improvement can be attributed to VADF allocating more inference steps to manipulation-critical stages identified by HVTS, which improves trajectory consistency and reduces stochastic sampling noise.

\section{Implementation Details of HVTS}
\label{sec:hvts_details}
\subsection{Setup}

All simulation experiments were conducted on a workstation with two NVIDIA RTX A6000 GPUs (48GB VRAM each). The implementation is based on the Hugging Face \texttt{transformers} library (v4.45.0 or later) without Flash Attention 2 to maintain a vanilla baseline.

\noindent \textbf{Model Configuration.}
We use \texttt{Qwen2-VL-7B-Instruct} as the vision-language backbone in BF16 precision. The VLM is hosted on \textbf{GPU 0}, while \textbf{GPU 1} executes the HVTS module to avoid contention between visual reasoning and diffusion policy inference.

\noindent \textbf{Input Processing.}
Images are processed using \texttt{Qwen2VLProcessor} with dynamic resolution scaling. The pixel range is constrained to

\begin{equation}
P_{min} = 256 \times 28 \times 28,
\quad
P_{max} = 1280 \times 28 \times 28
\end{equation}

to balance visual fidelity and memory usage.

Before being passed to the VLM, images are converted to RGB, resized while preserving aspect ratio, and center-padded to a square resolution of $448 \times 448$.

\noindent \textbf{Decoding Configuration.}
All VLM generations use the following decoding settings:

\begin{itemize}
\item \texttt{max\_new\_tokens}: 1024
\item \texttt{temperature}: 0.1
\item \texttt{top\_p}: 0.7
\item \texttt{padding}: True
\end{itemize}

\subsection{Task Decomposition: Stage Template Generation}
\label{subsec:task_decomp}

HVTS first decomposes a long-horizon task into semantic stages using a zero-shot VLM. We feed a sequence of uniformly sampled keyframes from an expert demonstration video into \texttt{Qwen2-VL-7B-Instruct}. The model identifies visually distinct manipulation phases and outputs a structured list of stage templates.

\noindent \textbf{Input Configuration.}
The model receives the sampled keyframes together with a task-specific prompt that enforces a fixed number of stages ($N$) and a structured JSON output. Stage decomposition is performed once per task using a single expert trajectory.

\begin{tcolorbox}[colback=gray!5,colframe=gray!60,title=Prompt for Stage Decomposition]

\textbf{Task:} [task\_desc]

You are given [len(processed\_images)] images showing the progression of the task. Decompose the task into exactly [num\_stages] stages based on visual changes. \\
Each stage should describe the pixel-level visual change between states. \\
Naming rule: task/stage names should use underscores instead of spaces \\

Return exactly [num\_stages] stages in JSON with schema:
\begin{verbatim}
[
  {
    "name": "<stage_name>",
    "description": "Action features: <desc>"
  }
]
\end{verbatim}

\end{tcolorbox}

The returned JSON is sanitized using lightweight post-processing (e.g., removing trailing commas or non-JSON tokens) before parsing.

\noindent
The \texttt{name} field defines the discrete stage space used by the controller, while the \texttt{description} field provides semantic cues used during stage scheduling.

Listing \ref{lst:vlm_output_short} shows an example output for the ``Can'' task.

\begin{lstlisting}[
    caption={Abridged VLM Output for Stage Templates},
    label={lst:vlm_output_short},
    basicstyle=\small\ttfamily,
    frame=single,
    breaklines=true,
    postbreak=\mbox{\textcolor{red}{$\hookrightarrow$}\space}
]
[
    {
        "name": "robot_arm_initial_position",
        "description": "Action features: The robot arm is positioned above the red can..."
    },
    {
        "name": "robot_arm_grasps_can",
        "description": "Action features: The robot arm's gripper closes around the red can..."
    },
    {
        "name": "robot_arm_moves_toward_container",
        "description": "Action features: The robot arm moves horizontally toward the container..."
    },
    {
        "name": "robot_arm_releases_can_into_compartment",
        "description": "Action features: The robot arm lowers the red can into the compartment..."
    },
    {
        "name": "robot_arm_post_placement",
        "description": "Action features: The robot arm retracts slightly after..."
    }
]
\end{lstlisting}

The set of stage names defined in Listing \ref{lst:vlm_output_short} is subsequently mapped to low-level control parameters, as detailed in Section \ref{subsec:step_scheduling}.

\noindent \subsection{Dynamic Scheduling}
\label{subsec:step_scheduling}

\begin{tcolorbox}[colback=gray!5,colframe=gray!60,title=Prompt for Stage Parameter Assignment]

Task stages (total [num stages]):[stage definitions]\\
Assign two parameters for each stage:\\
- n\_action\_steps: integer in $[a_{min}, a_{max}]$ \\
- num\_inference\_steps: integer in $[i_{min}, i_{max}]$ \\
Choose n\_action\_steps and num\_inference\_steps based on the relative difficulty of each stage. \\
Use smaller values for simple stages and larger values for more precise stages. \\
Do not assign the same values to all stages.\\

Return JSON for all stages:\\
\begin{verbatim}
[
  {
    "name": "<stage_name>",
    "n_action_steps": <N_a>,
    "num_inference_steps": <N_d>
  }
]
\end{verbatim}

\end{tcolorbox}

After generating stage templates, HVTS assigns each semantic stage a diffusion inference schedule. This module converts the task decomposition into a stage-conditioned parameter table used during policy execution.

\noindent \textbf{Input Configuration.}
The input consists of the stage definitions produced in Section~\ref{subsec:task_decomp}, including stage names and textual descriptions. Given these definitions together with the task description, the VLM assigns two inference parameters to each stage: the action horizon $N_a$ and the number of denoising steps $N_d$.

To ensure that at least one precision-critical phase receives higher computation, the model is required to designate exactly one stage as the most difficult stage and assign it the minimum $N_a$ and maximum $N_d$ within predefined ranges.

The same \texttt{Qwen2-VL-7B-Instruct} model is used for this stage.

\noindent \textbf{Output Format.}
The model returns a schedule table that maps each stage to a parameter pair $(N_a^k, N_d^k)$. The table is generated offline and used as a control prior during policy execution.

Listing \ref{lst:schedule_output_short} shows an example schedule for the ``Can'' task.

\begin{lstlisting}[
    caption={Abridged Output for Stage-to-Schedule Assignment},
    label={lst:schedule_output_short},
    basicstyle=\small\ttfamily,
    frame=single,
    breaklines=true,
    postbreak=\mbox{\textcolor{red}{$\hookrightarrow$}\space}
]
[
    {
        "name": "robot_arm_initial_position",
        "n_action_steps": 16,
        "num_inference_steps": 20
    },
    {
        "name": "robot_arm_grasps_can",
        "n_action_steps": 16,
        "num_inference_steps": 40
    },
    {
        "name": "robot_arm_moves_toward_container",
        "n_action_steps": 16,
        "num_inference_steps": 40
    },
    {
        "name": "robot_arm_releases_can_into_compartment",
        "n_action_steps": 8,
        "num_inference_steps": 60
    },
    {
        "name": "robot_arm_post_placement",
        "n_action_steps": 16,
        "num_inference_steps": 20
    }
]
\end{lstlisting}

\noindent \textbf{Schedule Table Construction.}
The generated output is stored as a stage-conditioned schedule table, which maps each semantic stage to its corresponding inference parameters. During execution, the controller queries this table to retrieve the corresponding $(N_a, N_d)$ pair once the current manipulation stage is identified.If the generated schedule does not explicitly include a designated hardest stage, a lightweight heuristic fallback is applied to enforce one high-computation precision stage.

\subsection{Stage Classification: Online Inference}
\label{subsec:stage_classification}

During policy execution, HVTS performs online stage classification to determine the current manipulation phase and retrieve the corresponding inference schedule $(N_a, N_d)$ from the schedule table described in Section~\ref{subsec:step_scheduling}.

\noindent \textbf{Input Configuration.}
At runtime, the system maintains a short temporal buffer of recent RGB observations. A sequence of consecutive frames is provided to the VLM to capture the visual progression of the manipulation process. These frames are ordered chronologically from earliest to most recent, allowing the model to infer the current stage by analyzing both motion cues and the latest visual evidence. The stage templates generated in Section~\ref{subsec:task_decomp} are also supplied as semantic references.

\begin{tcolorbox}[colback=gray!5,colframe=gray!60,title=Prompt for Stage Classification]

Task: You are given several consecutive frames from a robotic manipulation task.  
The images are ordered chronologically from earliest to most recent.

Analyze the visual progression and determine the \textbf{current stage} of the task.
Focus primarily on the most recent frame while considering the temporal evolution.

Stages:
[stage definitions]

Return the top-$k$ most likely stages ranked by probability.

Output format:

\begin{verbatim}
stage_name: probability
stage_name: probability
stage_name: probability
\end{verbatim}

Only output the stage names and probabilities without additional explanations.

\end{tcolorbox}

\noindent \textbf{Output Format.}
The VLM returns the top-$k$ candidate stages together with confidence scores. These probabilities represent the model's relative belief over possible manipulation phases given the current observation sequence.

\noindent \textbf{Inference Procedure.}
To reduce computational overhead, stage classification is not executed at every timestep. Instead, the system performs classification periodically after a fixed number of control steps. Between two classification events, the previously predicted stage is cached and reused.

When a classification event is triggered, the VLM produces a ranked list of candidate stages with probabilities. The final stage prediction is selected using the following rule:

\begin{itemize}
\item If the confidence gap between the top two candidates is sufficiently large, the most probable stage is selected.
\item If the probabilities are close, the final stage is sampled from the top-$k$ candidates according to their probabilities to avoid unstable switching near stage boundaries.
\end{itemize}

\noindent \textbf{Schedule Retrieval.}
Once the current stage $k$ is determined, the controller queries the precomputed schedule table to obtain the corresponding inference parameters $(N_a^k, N_d^k)$. These parameters are then used for the subsequent diffusion policy inference steps until the next classification event occurs.

\end{document}